# Intelligent Optimization of Multi-Parameter Micromixers Using a Scientific Machine Learning Framework

Meraj Hassanzadeh, Ehsan Ghaderi, Mohamad Ali Bijarchi[1], and Siamak Kazemzadeh Hannani

Department of Mechanical Engineering, Sharif University of Technology, Tehran, Iran

**Abstract**

Multidimensional optimization has consistently been a critical challenge in engineering. However, traditional simulation-based optimization methods have long been plagued by significant limitations: they are typically capable of optimizing only a single problem at a time and require substantial computational time for meshing and numerical simulation. This paper introduces a novel framework leveraging cutting-edge Scientific Machine Learning (Sci-ML) methodologies to overcome these inherent drawbacks of conventional approaches. The proposed method provides instantaneous solutions to a spectrum of complex, multidimensional optimization problems. A micromixer case study is employed to demonstrate this methodology. An agent, operating on a Deep Reinforcement Learning (DRL) architecture, serves as the optimizer to explore the relationships between key problem parameters. This optimizer interacts with an environment constituted by a parametric Physics-Informed Neural Network (PINN), which responds to the agent's actions at a significantly higher speed than traditional numerical methods. The agent's objective, conditioned on the Schmidt number is to discover the optimal geometric and physical parameters that maximize the micromixer's efficiency. After training the agent across a wide range of Schmidt numbers, we analyzed the resulting optimal designs. Across this entire spectrum, the achieved efficiency was consistently greater than the baseline, normalized value. The maximum efficiency occurred at a Schmidt number of 13.3, demonstrating an improvement of approximately 32%. Finally, a comparative analysis with a Genetic Algorithm was conducted under equivalent conditions to underscore the advantages of the proposed method.

---

[1] corresponding author: bijarchi@sharif.edu

# 1. Introduction

Recently, the use of micromixers in various microfluidic applications has attracted considerable attention from researchers [1]. These systems, due to their compact size, fast response time, and low cost, are ideal candidates for lab-on-a-chip studies [2]. Generally, micromixers are categorized into two main types: active and passive [3]. Active micromixers promote fluid mixing by utilizing external energy sources such as electric, magnetic, or ultrasonic fields [4-5]. However, the requirement for external energy and the associated high costs limit the practical applications of active methods [6]. In contrast, passive micromixers rely solely on carefully designed channel geometries to disrupt the fluid flow and enhance mixing efficiency [7]. Given the low fluid flow velocities at the microscale, mixing processes occur slowly, which has driven researchers to continuously seek effective strategies for improving mixing performance [8].

One of the important passive methods in micromixer applications is the use of baffles within the microchannel. These internal structures disturb the flow behavior and generate vortices, thereby enhancing fluid mixing and increasing key performance indicators such as the Mixing Index (MI) and Mixing Efficiency (ME) [9-10]. The integration of baffles with various shapes in both two-dimensional and three-dimensional microchannels has been a focus of recent studies [11-12]. Three-dimensional investigations of this concept have been conducted by researchers such as Afzal and Kim [13], and Chen and Zhao [14], while Chen and Lv [15] explored the effects of the ratio between three geometric parameters of the micromixer in 3D microchannels containing internal baffles. Due to the complexity and high computational cost associated with simulating three-dimensional micromixers, many researchers have turned to two-dimensional analyses as an efficient alternative for studying the mixing behavior in such systems [16-18]. Various experimental and numerical methods have been employed by researchers to study the phenomenon of fluid mixing [19-20].

In addition to traditional approaches, recent studies have increasingly focused on the use of Artificial Intelligence (AI) and Machine Learning (ML) techniques to predict fluid behavior [21]. Despite recent advancements in AI and data-driven modeling, the limited availability of experimental data and the relatively low accuracy of numerical methods in capturing all possible flow conditions have posed significant challenges [22]. To address these limitations, a new class of approaches known as Scientific Machine Learning (SciML) has emerged [23]. SciML aims to integrate existing data with physical knowledge of the system to simulate and predict complex fluid flow behavior. One of the key developments in this field has been introduced by Raissi et al. [24] under the framework of Physics-Informed Neural Networks (PINNs). This method combines the strengths of AI—particularly deep neural networks—with the governing equations and boundary conditions of the physical system. Remarkably, PINNs have demonstrated the ability to predict fluid behavior even in the absence of training data, relying solely on the embedded physics of the problem [25].

In recent years, researchers have utilized the PINN method to investigate fluid flow behavior across various engineering applications [26-28]. Specifically, in the field of fluid mixing and mass transfer, limited studies have been conducted, such as the work by Kou et al. [29], who investigated the integration of the mass transfer equation in mixing phenomena. The fluid concentration distribution was predicted in addition to a limited amount of experimental data. Additionally, Hassanzadeh et al. [30] studied a three-dimensional micromixer equipped with baffles of various shapes and configurations under different Reynolds numbers. The adopted flexible network architecture successfully handled the problem's complexity, yielding accurate and reliable predictions of flow behavior. Owing to the multitude of physical and geometric parameters characterizing the problem, parametric investigation using PINNs has been adopted by researchers. In the realm of parametric modeling using PINN, Beltrán-Pulido et al. [31] examined two-dimensional electro-magnetic flow problems. Various magnetic field parameters were introduced as network inputs, enabling the model to predict flow behavior under different physical conditions. The results indicate a maximum absolute error of 2.2%. In another study, Arthurs and King [32] treated the geometry as a parametric input to the neural network, applying this approach to channel walls of varying shapes. The simulation was conducted parametrically to capture the effect of geometric variations. Furthermore, Ghaderi et al. [33] investigated fluid flow within channels with various radius of obstruction under magnetic fields, incorporating multiple geometric and physical parameters as inputs to the learning process. This enabled the model to generalize effectively across different scenarios. Parametric modeling has also attracted attention in other applications such as high Reynolds number flows [34], turbulent flow simulations [35], and combustion problems [36], demonstrating its broad utility in complex fluid dynamics challenges.

Parametric studies using the PINN method are often employed in optimization problems, as a single solution of the problem enables prediction of the output for various network input values. This capability can significantly accelerate the solution process in optimization tasks [37-38]. For example, Sun et al. [39] employed a parametric solution for the flow around an airfoil to determine the optimal angle of attack. The optimization algorithm used during the network training phase of the PINN method was also applied to optimize the angle of attack at the network input stage. In another study, Priyadarshi et al. [40] incorporated the physics of the problem into the network inputs and performed the optimization after completing the PINN process, using the data obtained from the trained network. Owing to the solutions provided by the PINN method, the optimization stage could be conducted on the resulting dataset in a short computational time. Recently, the use of advanced machine learning algorithms—such as Reinforcement Learning (RL)—has garnered attention in optimization [41-42]. In RL, the goal is to optimize a target parameter through interaction with the environment and a trial-and-error process, adapting to changes in other parameters [43]. While the application of RL in PINN-based learning processes has been investigated by Banerjee et al. [44], to the best of our knowledge, its use for optimization within the PINN framework has not yet been reported in the literature.

Based on the points mentioned above, no study to date has utilized RL to optimize solutions obtained from parametric studies conducted via PINNs. This study employs a Generalized Deep Reinforcement Learning (GDRL) framework for optimization. The primary objective is to develop a model that does not require retraining for different physical conditions, such as varying Schmidt or Reynolds numbers. The key advantage of this method is its generalizability across a wide range of physical problems. Unlike traditional optimization techniques, which are typically limited to finding an optimal solution for a single specific problem, the proposed GDRL model, once trained, can instantly provide optimal solutions for a broad spectrum of physical scenarios. To evaluate the method's performance, a two-dimensional micromixer was selected as a case study. This problem is particularly suitable for demonstrating the method's capability in high-dimensional optimization, as it simultaneously depends on both geometric and physical parameters. The system is simulated using a parametrically trained PINN. The use of a PINN enables the concurrent execution of a large number of simulations without the need for retesting, thereby allowing for a more accurate and efficient comparison of different optimization strategies. Within this framework, the PINN model serves as the DRL environment, facilitating action deployment and feedback collection at a significantly higher speed than conventional numerical methods.

## 2. Problem Description

The geometry of the problem is illustrated in Figure 1. The shaded region represents the various possible configurations that may arise in the problem. Considering the parametric solution for different geometries, several sample spline profiles are presented in figure. Furthermore, a summary of the geometric dimensions is provided in Table 1.

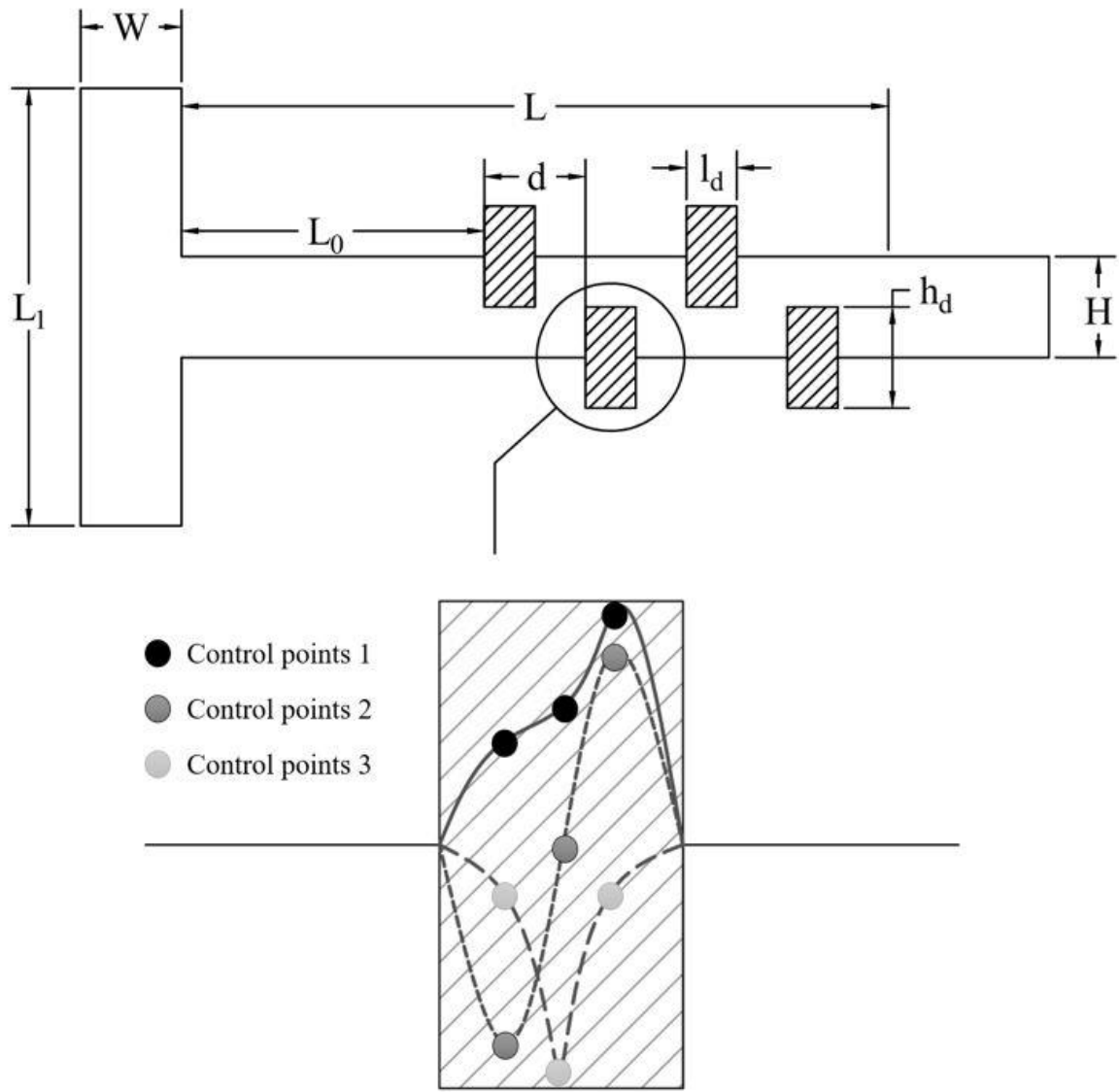

**Figure 1.** Schematic of the problem geometry and sample Spline profiles.

**Table 1:** Geometric specifications of the microchannel.

| Notation | Definition | Value |
|---|---|---|
| L | The Length of the microchannel | 2.1 mm |
| $L_0$ | First Spline inlet distance | 0.9 mm |
| $L_1$ | The length of the inlet of the microchannel | 1.3 mm |
| H | The height of the microchannel | 0.3 mm |
| W | The width of the microchannel inlet | 0.3 mm |
| d | Distance between two consecutive Spline | 0.15 mm |
| $h_d$ | Allowable height deviation for Spline | 0.3 mm |
| $l_d$ | Allowable length deviation for Spline | 0.15 mm |

The fluid flow enters the channel from the top and bottom, and the mixing phenomenon within the channel is investigated. The governing equations of the problem are expressed in vector form as Equations under the assumptions of incompressible and steady-state flow in two dimensions (1-3) [30].

$$\nabla . \vec{U} = 0 \tag{1}$$
$$\rho(\vec{U}.\nabla)\vec{U} = -\vec{\nabla}p + \mu \nabla^2 \vec{U} \tag{2}$$
$$(\vec{U}.\nabla)c = D\nabla^2 c \tag{3}$$

The governing equations—comprising the continuity, momentum, and mass transfer equations—involve first and second-order spatial derivatives. To streamline the solution process, these equations are reformulated into first-order derivatives [30]. This approach ensures that back propagation in the neural network is performed only once, enhancing computational efficiency. Accordingly, in this study, the equations are expressed in terms of lower-order, dimensionless derivatives, as detailed in Equations (4)–(12) in Cartesian coordinates. Furthermore, to improve solution accuracy, the dimensionless form of the equations is employed [30], which helps balance the terms in the loss function. Specifically, Equation (4) describes the continuity equation, Equations (5)–(6) represent the momentum equations in two spatial directions, and Equation (10) governs mass transfer. The remaining equations are constitutive relations that support the conservation equations. These equations are simplified into their dimensionless form, given by Equations (4-12) [30]. In these relations, $\vec{U}$ is the velocity vector, $u^*$ and $v^*$ are the dimensionless velocity components in the $x$ and $y$ directions, correspondingly. Additionally, $p^*$, $\tau^*$, and $c^*$ represent the dimensionless pressure, dimensionless stress, and dimensionless concentration, respectively. Moreover, $\rho$, $\mu$, $D$, and $U_m$ are the density, viscosity, diffusion coefficient of concentration, and the mean velocity of the fluid in the main channel, respectively. In these relations, $J^*$ is the dimensionless concentration flux, and the Reynolds and Schmidt numbers are defined in equations (13) and (14) [30].

$$\frac{\partial u^*}{\partial x^*} + \frac{\partial v^*}{\partial y^*} = 0 \tag{4}$$

$$\left(u^* \frac{\partial u^*}{\partial x^*} + v^* \frac{\partial u^*}{\partial y^*}\right) - \frac{\partial \tau^*_{xx}}{\partial x^*} - \frac{\partial \tau^*_{xy}}{\partial x^*} = 0 \tag{5}$$

$$\left(u^* \frac{\partial v^*}{\partial x^*} + v^* \frac{\partial v^*}{\partial y^*}\right) - \frac{\partial \tau^*_{xy}}{\partial y^*} - \frac{\partial \tau^*_{yy}}{\partial y^*} = 0 \tag{6}$$

$$-p^* + \frac{2}{\text{Re}} \frac{\partial u^*}{\partial x^*} - \tau^*_{xx} = 0 \tag{7}$$

$$-p^* + \frac{2}{\text{Re}} \frac{\partial v^*}{\partial y^*} - \tau^*_{yy} = 0 \tag{8}$$

$$\frac{1}{\text{Re}} \left(\frac{\partial u^*}{\partial y^*} + \frac{\partial v^*}{\partial x^*}\right) - \tau^*_{xy} = 0 \tag{9}$$

$$\left(u^* \frac{\partial c^*}{\partial x^*} + v^* \frac{\partial c^*}{\partial y^*}\right) + \frac{1}{\text{ReSc}} \left(\frac{\partial J^*_x}{\partial x^*} + \frac{\partial J^*_y}{\partial y^*}\right) = 0 \tag{10}$$

$$J^*_x + \frac{\partial c^*}{\partial x^*} = 0 \tag{11}$$

$$J^*_y + \frac{\partial c^*}{\partial y^*} = 0 \tag{12}$$

$$\text{Re} = \frac{\rho U_m H}{\mu} \tag{13}$$

$$\text{Sc} = \frac{\mu}{\rho D} \tag{14}$$

The above equations, along with the boundary conditions governing the mixing problem (as illustrated in Figure 2), are applied to the problem. As illustrated in the figure, a parabolic velocity profile is prescribed at the inlet, while different concentration boundary conditions are applied along the upper and lower channel inlets. Moreover, a no-slip boundary condition is imposed on all channel walls, including those defined by splines that act as baffles. Due to the geometric complexity of the spline-based walls, enforcing the zero mass flux boundary condition on these surfaces presents a significant challenge. One of the key contributions of this study lies in the high flexibility of the proposed method in handling such complex boundary conditions. For details on the spline formulation and its normal vectors, see Supplementary Section 1. The implementation is available in the GitHub repository.

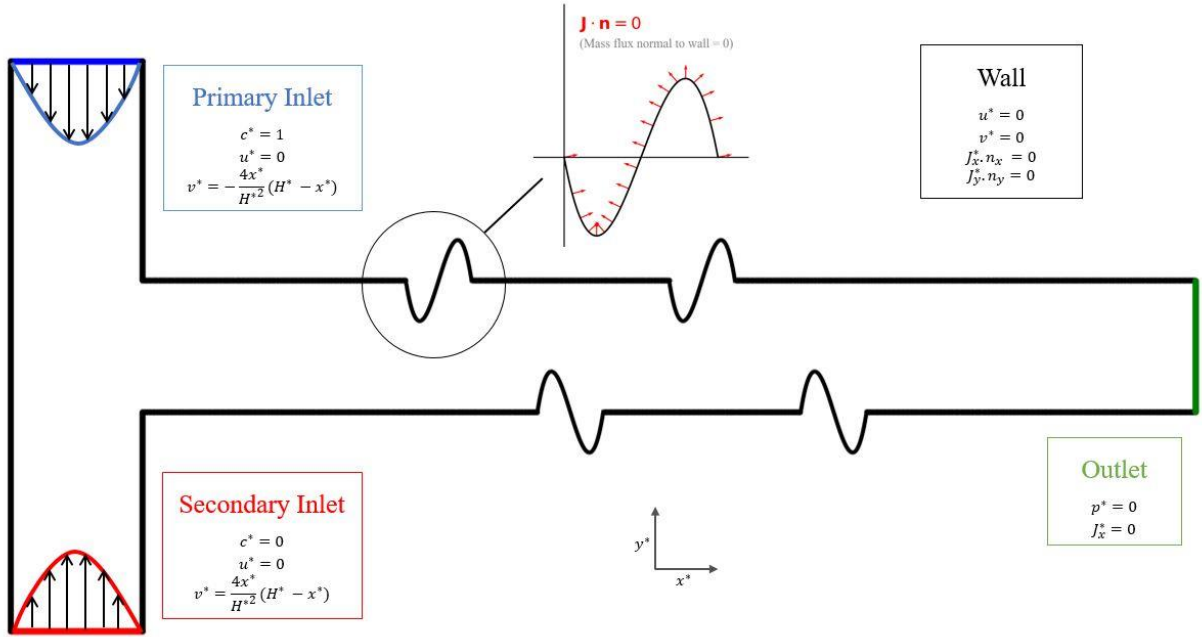

**Figure 2.** The boundary conditions of the problem.

## 3. Solution Methodology

This study introduces a framework that integrates a parametric Physics-Informed Neural Network (FlexPINN) with a Reinforcement Learning (RL) optimizer to automate the design of micromixers. The overall process, illustrated in Figure 3, consists of two main components: a pre-trained simulation environment and an RL-based design agent.

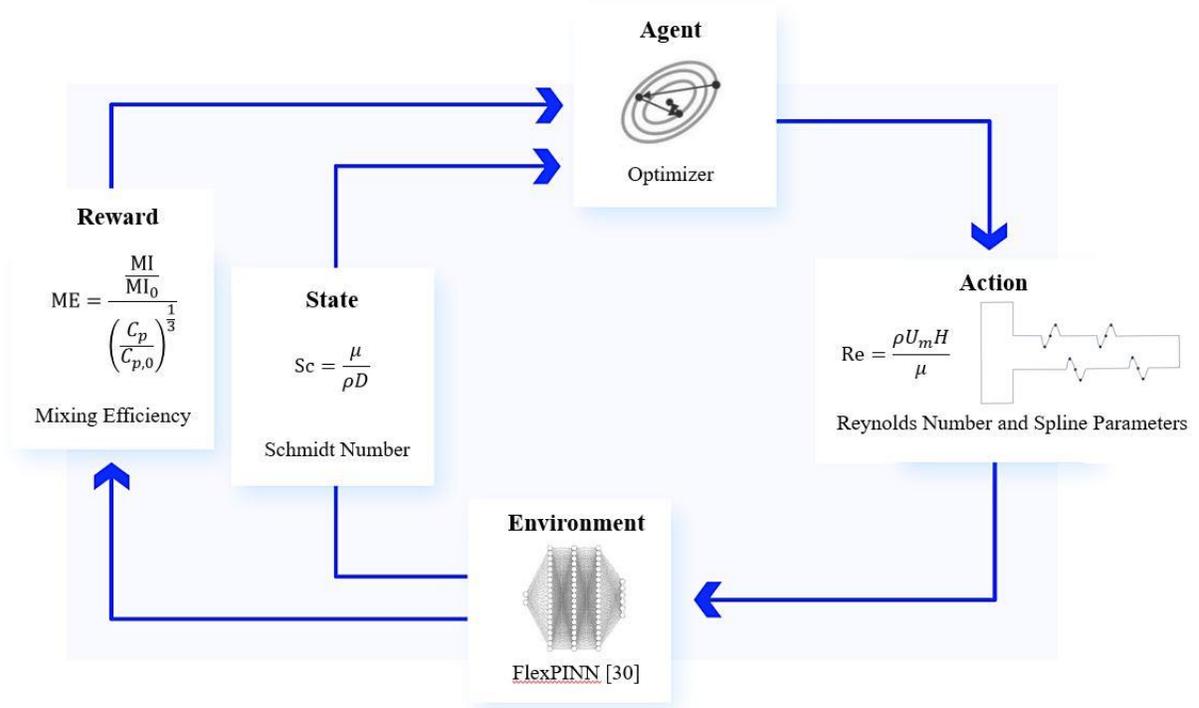

**Figure 3.** The architecture employed in this study.

## 3.1. The FlexPINN as a Simulation Environment

A FlexPINN model, extending the work of Hassanzadeh et al. [30], serves as the digital environment. This network takes spatial coordinates (x, y), physical parameters (Schmidt number, Reynolds number), and geometric parameters (spline shape) as inputs and outputs the resulting flow and concentration fields. By embedding the physics directly into the neural network's loss function, it acts as a fast and differentiable simulator, replacing computationally expensive numerical solvers. The complete implementation of the FlexPINN architecture and its training code are available in this work's [GitHub](#) repository.

## 3.2. Reinforcement Learning for Design Optimization

The core of the design automation is a Proximal Policy Optimization (PPO) agent [45], which learns to propose optimal designs through iterative interaction with the FlexPINN environment.

- **State (s):** The Schmidt number (Sc), representing the fluid properties.

- **Action (a):** A four-dimensional vector comprising the Reynolds number (Re) and three parameters defining the mixer's spline geometry.
- **Reward (r):** The Mixing Efficiency (ME), calculated from the FlexPINN-predicted concentration field according to Eq. (28) [30]. This reward incentivizes the agent to find designs that maximize mixing performance.

The agent's objective is to learn a policy ($\pi(a|s)$) that maps any given Schmidt number to the optimal set of physical and geometric parameters. The training involves a continuous loop: the agent proposes a design (action), the FlexPINN environment evaluates it and returns a mixing efficiency (reward), and the agent updates its policy to favor higher-rewarding actions (Algorithm 1).

$$\mathrm{ME} = \frac{\frac{\mathrm{MI}}{\mathrm{MI}_0}}{\left(\frac{C_p}{C_{p,0}}\right)^{\frac{1}{3}}} \tag{28}$$

$$C_p = \frac{\Delta P}{\frac{1}{2}\rho U_m^2} \tag{29}$$

$$\mathrm{MI} = 1 - \sqrt{\frac{1}{N}\sum_{i=1}^{N}\left(\frac{c_i^* - c_{mean}^*}{c_{mean}^*}\right)^2} \tag{30}$$

### 3.3. Training and Implementation

The PPO algorithm ensures stable learning through an actor-critic architecture and a clipped objective function. The actor network proposes actions, while the critic network evaluates the expected cumulative reward. The policy is updated by maximizing the objective $L^{CLIP(\theta)} = \mathbb{E}\left[\min\left(r_t(\theta)\hat{A}_t, clip(r_t(\theta), 1-\varepsilon, 1+\varepsilon)\hat{A}_t\right)\right]$, where $r_t(\theta)$ is the probability ratio between new and old policies, $\hat{A}_t$ is the advantage function estimated by the critic, and $\varepsilon$ is a clipping parameter that prevents overly large policy updates. For full algorithmic details, see Supplementary Section 2.

```
Algorithm 1 PPO for PINN-Based Mixing Optimization
 1: Initialize:
 2:   Actor π_θ, Critic V_φ networks
 3:   Action bounds: CP ∈ [−0.5, 0.5], Re ∈ [5, 40]
 4:   Hyperparams: γ = 0.99, ε = 0.2, K = 10
 5: for episode = 1 to E do
 6:     Sample batch: Sc ~ U(1, 100)
 7:     states ← [Sc_1, Sc_2, ..., Sc_N]
 8:     Actor Forward Pass:
 9:     μ, σ ← π_θ(states)
10:     actions ~ N(μ, σ)
11:     Scale actions to physical bounds
12:     PINN Evaluation:
13:     for each (action, state) pair do
14:         Generate mesh points for PINN
15:         c ← PINN(outlet)                              ▷ Concentration
16:         p ← PINN(inlet)                               ▷ Pressure
17:         MI ← 1 − √(Σ((c−0.5)/0.5)² / N)
18:         reward ← (MI/MI_0)/(p/p_0)^(1/3)
19:     end for
20:     PPO Update:
21:     advantages ← rewards − V_φ(states)
22:     Normalize advantages
23:     for k = 1 to K do
24:         π_{θ_new} ← π_θ(states)
25:         ratio ← π_{θ_new} / π_{θ_old}
26:         L^CLIP ← E[min(ratio · A, clip(ratio, 1−ε, 1+ε) · A)]
27:         L^VF ← (rewards − V_φ(states))²
28:         Update θ, φ using L^CLIP + L^VF
29:     end for
30: end for
```

To perform the simulations using the proposed method, the point distribution within the computational domain was generated via the Latin Hypercube Sampling (LHS) technique [46], as illustrated in Figure 4. Due to the parametric definition of the spline geometries in this study, certain points within these regions overlap in the figure. Furthermore, as outlined in the FlexPINN methodology, additional points were incorporated to account for the penalty term during the solution process. The source codes for geometry generation and the proposed solution method are provided at the following GitHub repository.

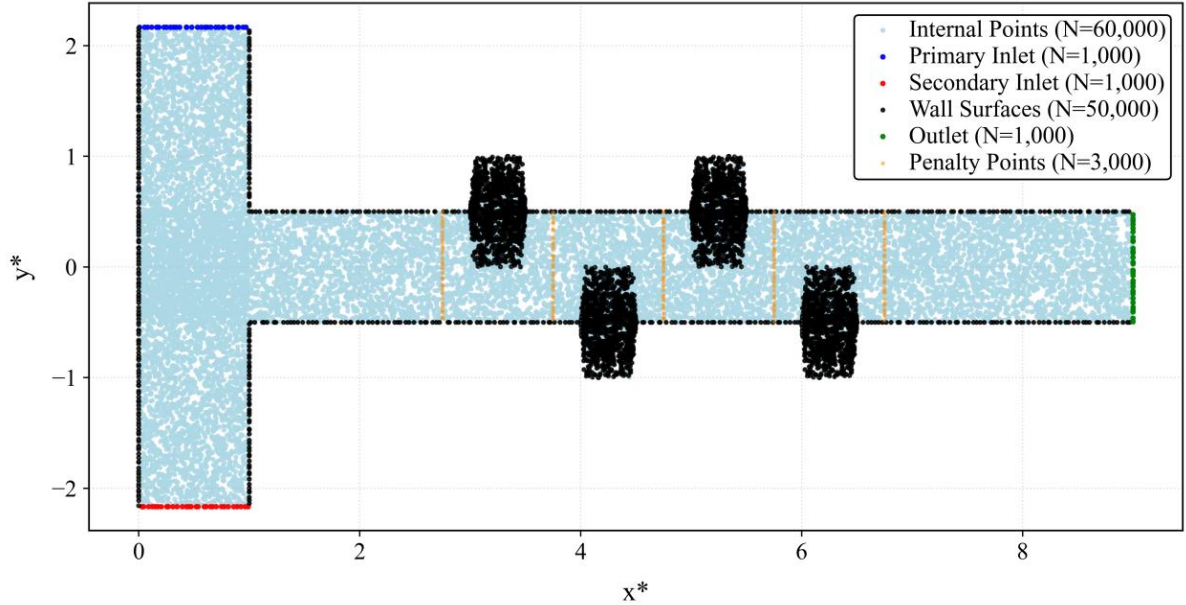

**Figure 4.** Point distributions generated using Latin Hypercube Sampling (LHS). (**Note:** The orange slices indicate regions where a constant mass flow rate was enforced as a penalty term in the FlexPINN method [30]).

## 4. Results and Discussion

The results derived from the described methodology are presented and discussed in this section. Figure 5 depicts the distributions of non-dimensional velocity, pressure, concentration, and the corresponding streamlines inside the channel. The investigation covered the following parameter ranges: Reynolds number, 5 to 40; Schmidt number, 1 to 100; and spline parameters, -0.5 to 0.5. Given the extensive parameter space defined by these geometric and physical variables, a randomized sampling approach was employed to select specific values for analysis. These results serve to confirm that the proposed algorithm produces physically realistic solutions.

a) 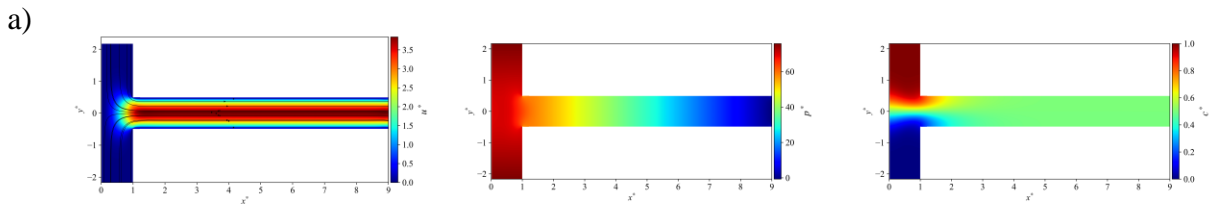

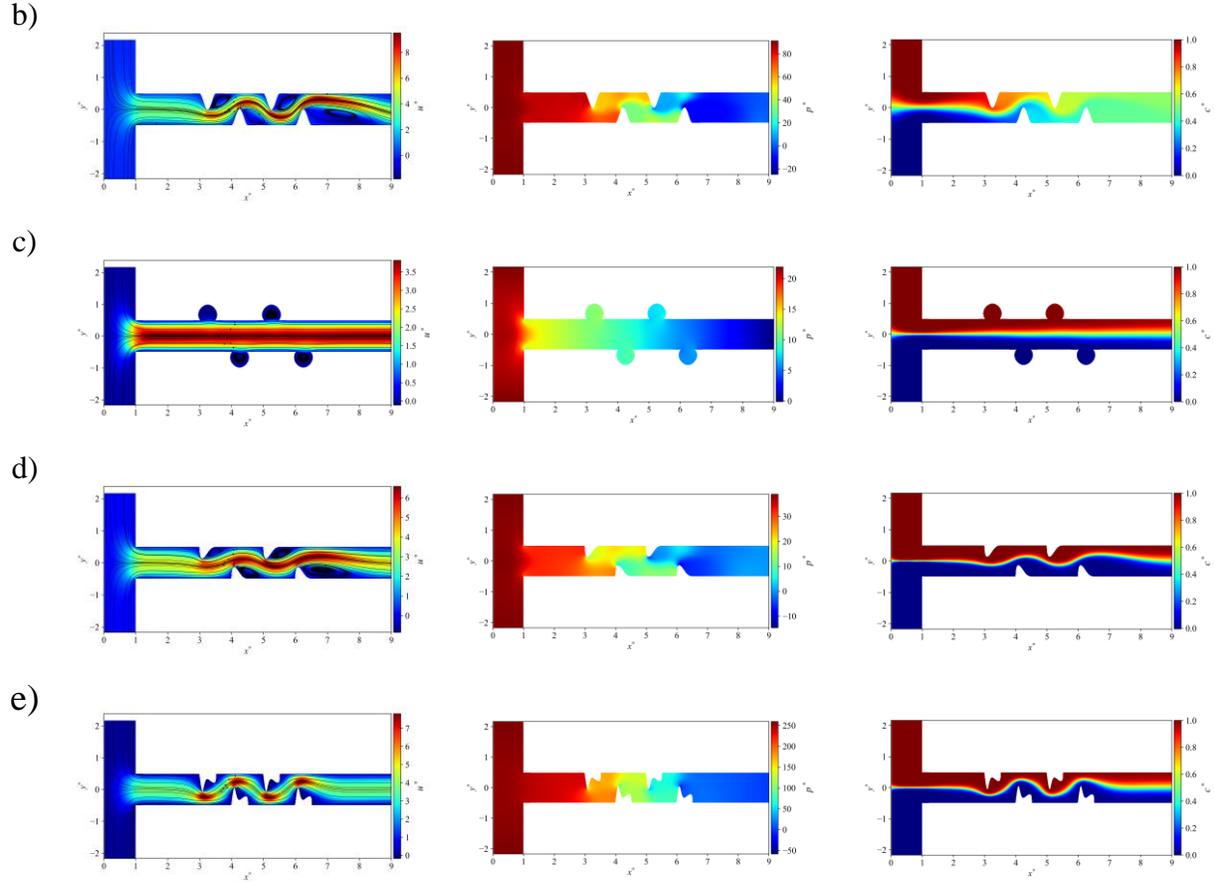

**Figure 5.** Contours of dimensionless velocity, pressure, and concentration (from left to right). a) Re = 5, CP1=0, CP2=0, CP3=0, Sc=1. b) Re =40, CP1=-0.25, CP2=-0.5, CP3=-0.25, Sc=1. c) Re =20, CP1=0.45, CP2=0.5, CP3=0.45, Sc=10. d) Re =40, CP1=-0.375, CP2=-0.1, CP3=-0.05, Sc=100. e) Re=5, CP1=-0.5, CP2=-0.1, CP3=-0.25, Sc=100.

## 4.1. Validation

The validation of the results was conducted by comparing the mixing efficiency, which serves as the key performance indicator in this analysis, with the findings reported by Shi et al. [47]. Their investigation, which employed a hybrid experimental-numerical approach, examined a range of channel and baffle geometries. From their study, two specific cases with geometries most analogous to the one used in the current research were chosen for comparison. Figure 6 demonstrates that the proposed method achieves excellent agreement, predicting the mixing efficiency with a maximum relative error under 4%. Additionally, it should be noted that in this figure (Figure 6(a)), the term "$ME_0$" refers to the baseline case of a channel without a baffle.

(a)

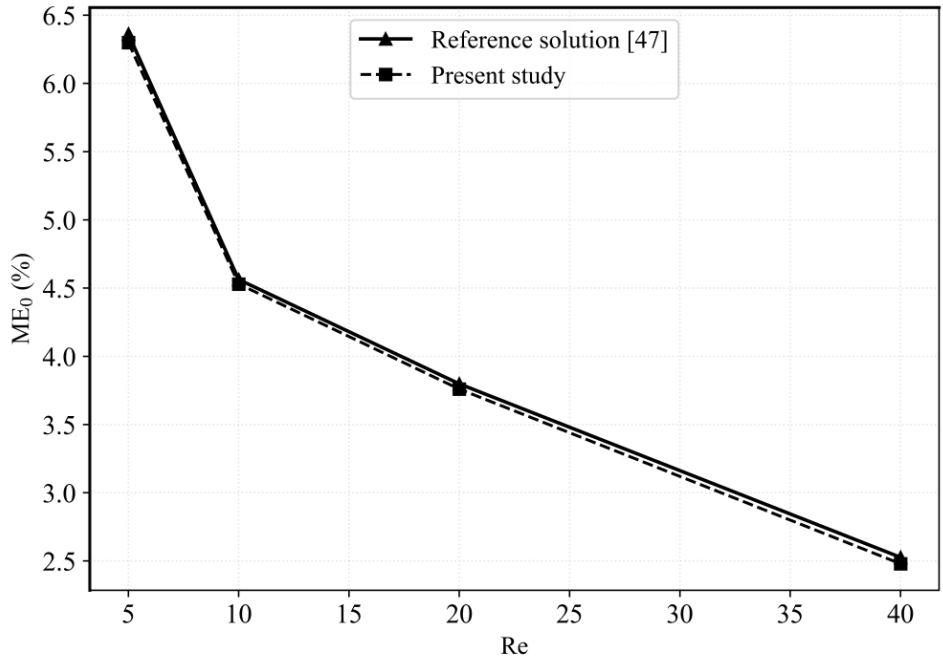

(b)

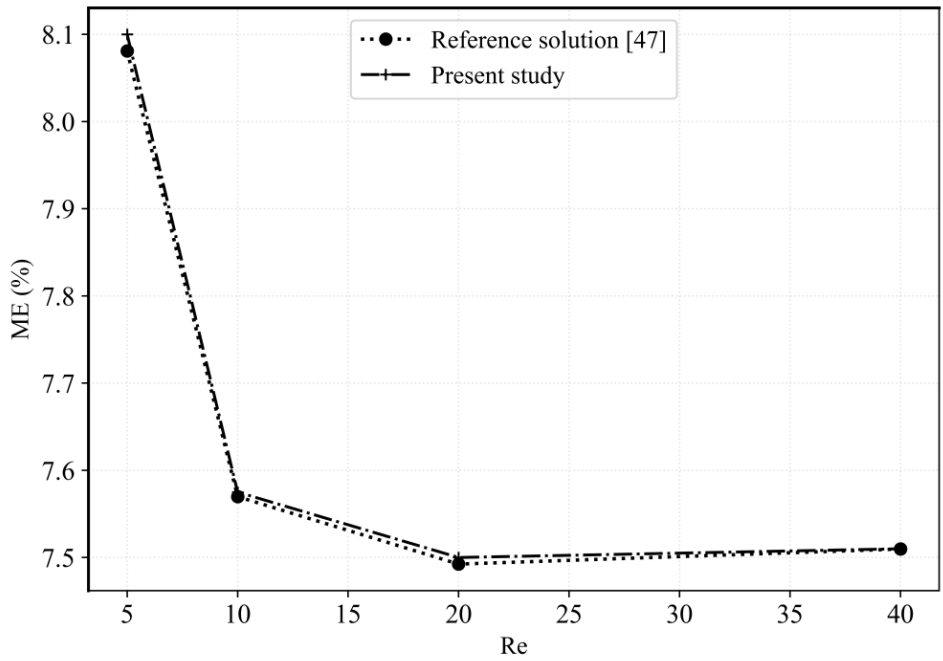

**Figure 6.** Comparison with the benchmark results of Shi et al. [47] for model validation: (a) channel without a baffle, (b) channel with a baffle.

## 4.2. The influence of parameters on the Mixing Efficiency (ME)

This section investigates the influence of various physical and geometric parameters on mixing efficiency using the proposed methodology. Initially, a channel without baffle (where all spline parameters are set to zero) is considered. The effects of the Reynolds and Schmidt numbers on the baseline mixing efficiency (denoted as $ME_0$) are presented in Figure 7a. The value of $ME_0$ is subsequently used to normalize mixing efficiency values in all other studied cases. As illustrated in the figure, the mixing efficiency in the channel without baffle increases with a decrease in both the Reynolds number and the Schmidt number. This behavior can be attributed to two primary factors. First, in the absence of a baffle, an increase in the Reynolds number leads to a significant pressure drop within the channel. The detrimental effect of this pressure loss on the flow regime outweighs its potential benefit for enhancing mass transfer, resulting in a net reduction of mixing efficiency. Second, an increase in the Schmidt number signifies a reduction in the relative importance of molecular diffusion, consequently, the mixing between the two fluids is diminished [48]. Although fluid advection may be enhanced at higher Reynolds numbers, the absence of a baffle means there is no mechanism present to capitalize on this by disrupting the flow and actively promoting mixing [49].

Moreover, in this work, the Schmidt number is treated as a core parameter, with other parameters analyzed relative to its influence. Accordingly, the Schmidt number is held constant at values of 1, 10, and 100 in Figures 7b, 7c, and 7d, respectively. These figures illustrate the combined effects of the Reynolds number and spline parameters on mixing efficiency under each condition. Figure 7b corresponds to a Schmidt number regime where molecular diffusion is dominant. Consistent with the trend observed in Figure 7a, the relative mixing efficiency in this regime decreases with an increasing Reynolds number. Furthermore, the results indicate that baffle configurations positioned externally and exhibiting symmetry yield superior performance compared to those that are internally positioned and asymmetric. This behavior can be interpreted as evidence that, under these conditions, the system's performance is more significantly influenced by pressure drop than by mass transfer enhancement. Consequently, the pressure loss plays a more dominant role in determining the mixing efficiency.

Figure 7c, corresponding to a Schmidt number of 10, reveals a non-monotonic trend in the relative mixing efficiency. For all baffle configurations, the efficiency initially decreases with increasing Reynolds number up to a certain threshold, beyond which a consistent increase is observed. This trend can be attributed to a transition in the dominant mixing mechanism. At intermediate Reynolds numbers, molecular diffusion does not have sufficient time to act, while advective transport is not yet strong enough to enhance mixing; consequently, the overall mixing efficiency decreases. In transitional Schmidt number regimes, the interplay between molecular diffusion and Reynolds-dependent convection leads to non-intuitive variations in mixing performance, such that the optimal condition occurs not at low or high Reynolds numbers, but near one of these extremes. At lower Reynolds numbers, molecular diffusion is the predominant factor. Beyond a critical Reynolds number, the influence of convective transport becomes more pronounced, with its effectiveness varying significantly based on the specific baffle geometry. This indicates that a Schmidt number of 10 lies within a transitional regime where the contributions of molecular diffusion and convective transport are comparable, leading to the distinct behavioral differences observed between internally and externally positioned baffles. Specifically, in the diffusion-dominated regime, the external baffles demonstrate superior performance, whereas the internal baffles exhibit better mixing efficiency when convective transport becomes the dominant mechanism. Furthermore, consistent with the other cases, the described behavior shows negligible dependence on the symmetry or asymmetry of the external baffles.

Figure 7d, corresponding to a Schmidt number of 100, demonstrates a consistent increase in relative mixing efficiency with the Reynolds number across all configurations. This trend is attributed to the dominant role of convective mass transfer over molecular diffusion at high Schmidt numbers. Under these conditions, an increase in the Reynolds number directly enhances the convective transport of concentration, thereby improving mixing efficiency. Furthermore, the results indicate that at low Reynolds numbers, the pressure drop has a more pronounced effect than mass transfer. Consequently, the presence of internal baffles in this regime does not lead to significant performance improvement. However, as the Reynolds number increases and convective effects intensify, the internal baffles induce a superior enhancement in mass transfer, resulting in a higher mixing efficiency compared to external baffles. Additionally, at high Reynolds numbers, the asymmetric internal baffles outperform their symmetric counterparts. This superior performance suggests that the gain in mass transfer achieved by the asymmetric geometry outweighs the associated increase in pressure drop.

a)

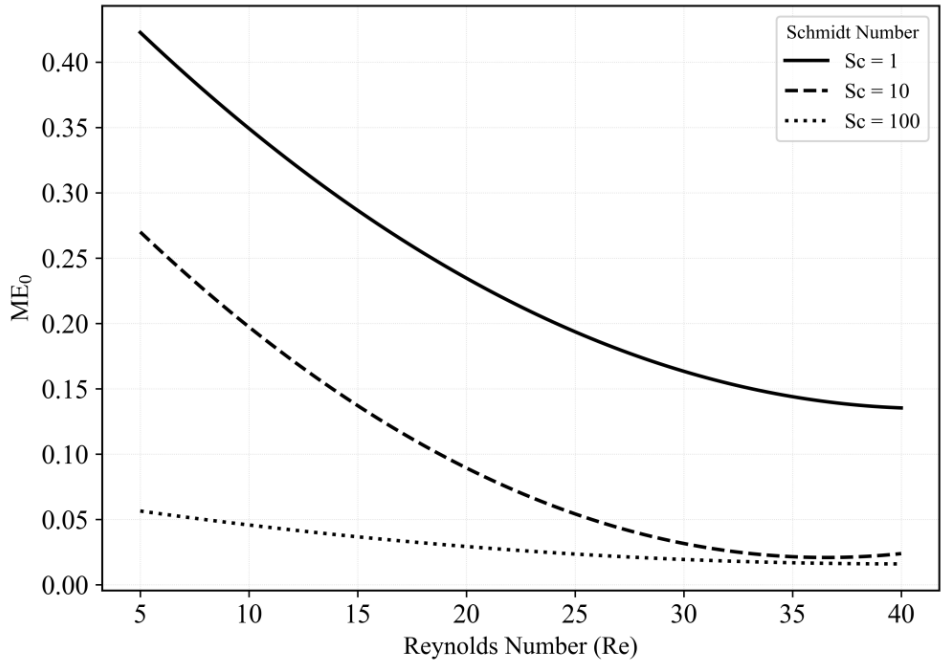

b)

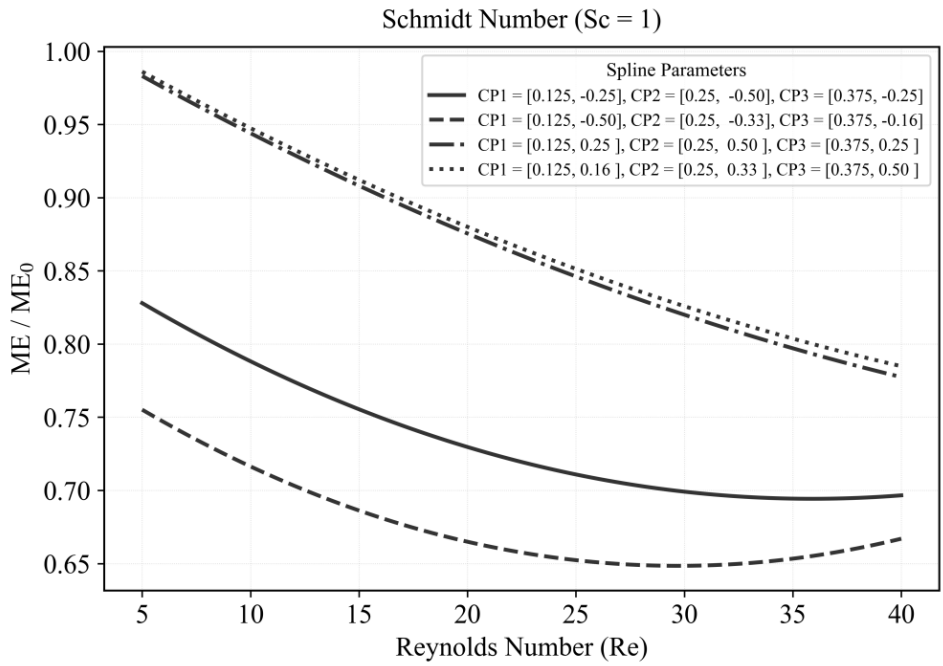

c)

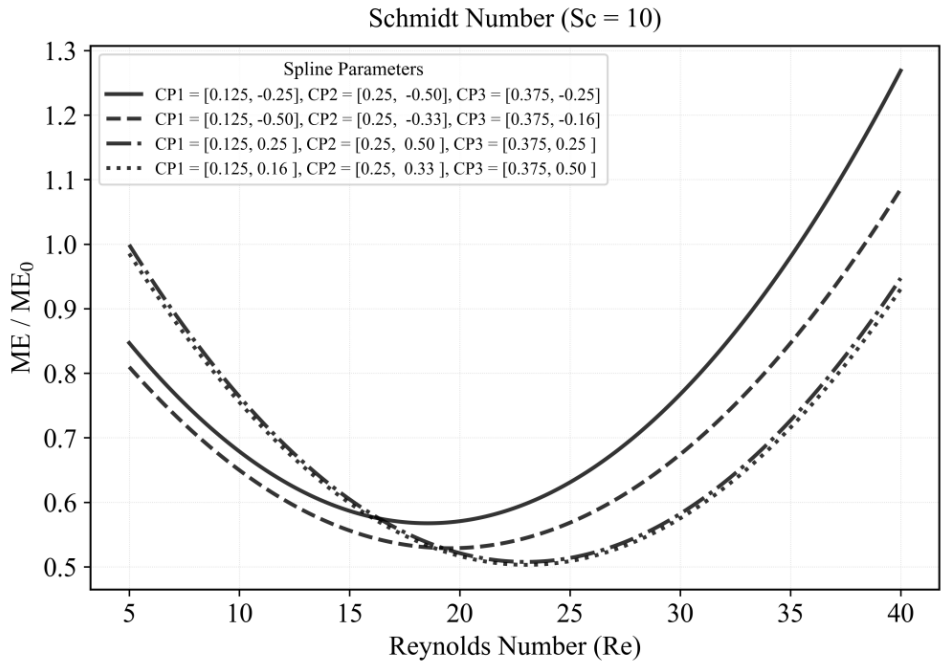

d)

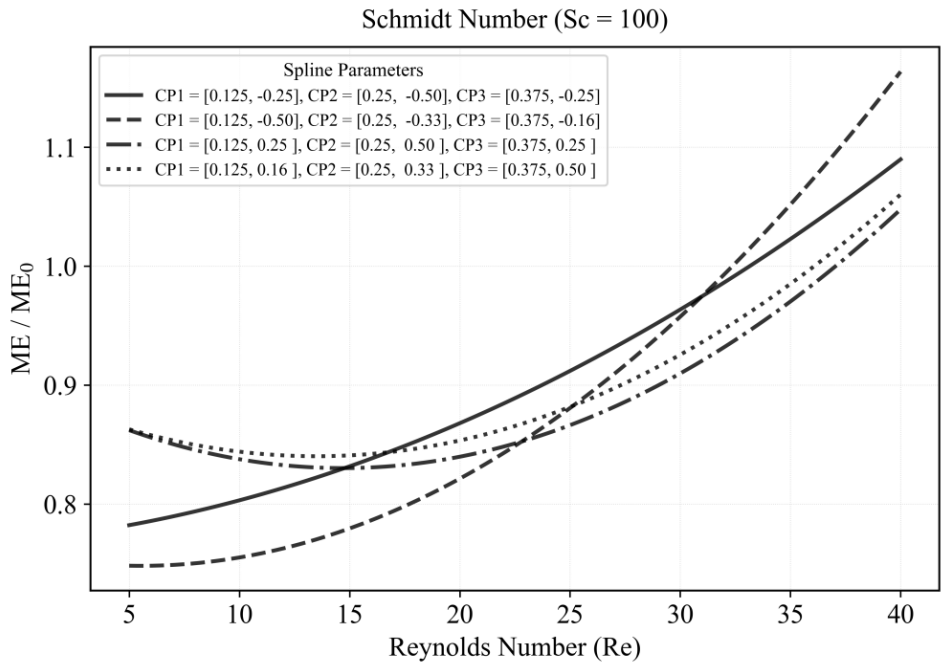

**Figure 7.** Influence of geometric and physical parameters on mixing efficiency. (a) Variation of the baseline mixing efficiency (ME$_0$) with Reynolds and Schmidt numbers. (b) Variation of mixing efficiency with Reynolds number and spline parameters at a Schmidt number of 1. (c) Variation of mixing efficiency with Reynolds number and spline parameters at a Schmidt

number of 10. (d) Variation of mixing efficiency with Reynolds number and spline parameters at a Schmidt number of 100.

## 4.3. Optimization

This section presents the optimization of the objective function, mixing efficiency, with respect to the physical and geometric parameters discussed in the previous section. Figure 8 illustrates the progression of the reward as a function of the epoch during the optimization process conducted via the Deep Reinforcement Learning (Deep RL) method. As the figure indicates, the fluctuations in the reward diminish over time, demonstrating the algorithm's successful convergence towards an optimal answer. In the figure, the "smoothed" curve represents the moving average of the reward, calculated over a window of 50 episodes. Given the extensive range of the geometric and physical parameters under investigation, the learning process exhibits significant initial oscillations. To clearly depict the underlying convergence trend amidst this fluctuation, the plot concurrently displays both the raw reward and its smoothed value.

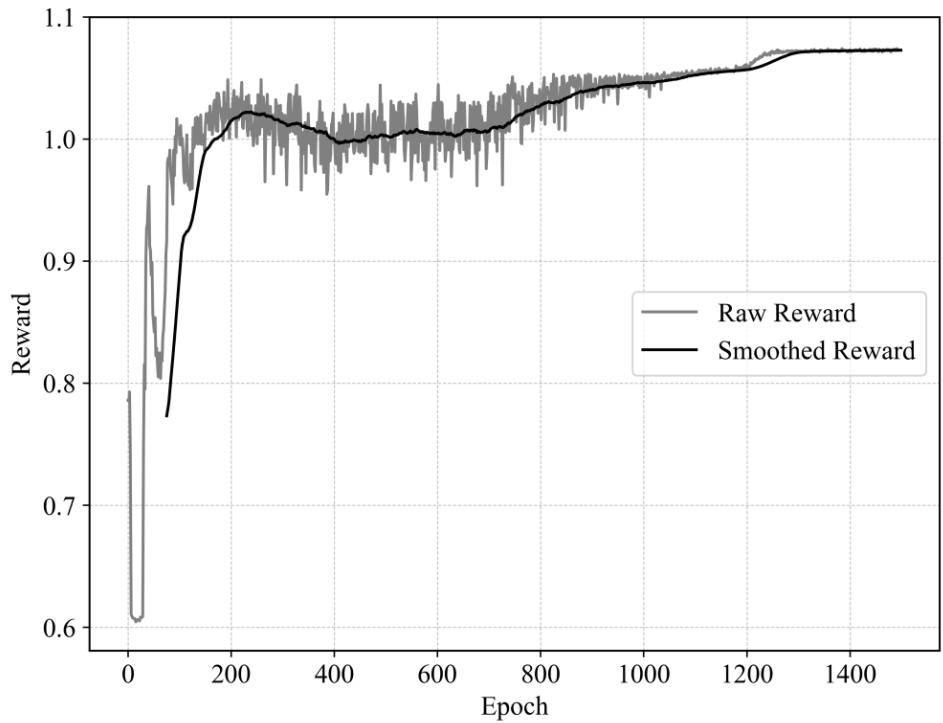

**Figure 8.** Reward progression during the optimization process using the PINN-RL algorithm.

Through the optimization process, the optimal configurations were extracted from the agent, and their corresponding relative mixing efficiencies for all Schmidt numbers are presented in Figure 9. The overall trend indicates an increase in relative efficiency with the Schmidt number up to a value of 13.3, beyond which the trend reverses and the efficiency decreases. For selected Schmidt numbers, the optimal parameters (Reynolds number and spline control points) are specified. Additionally, a cropped velocity contour plot around the baffle is provided for each case. The morphology of the optimal baffle evolved consistently with the dominant mixing mechanism: for Schmidt numbers up to 13.3, where diffusive mixing is dominant, the optimal configuration consistently featured a symmetric, internally-mounted baffle. At the transition point of Sc=13.3, this internal baffle reached its largest, most pronounced symmetric, streamlined shape. Beyond this critical point, as convective effects grew dominant, the optimal geometry shifted towards an asymmetric, internal configuration. This shift resulted in the formation of larger vortices behind the baffle. While this enhanced convective mixing, the associated increase in pressure loss was more significant, leading to the observed net reduction in overall relative mixing efficiency. Complete animated sequences of the results are available in the [GitHub](#) repository.

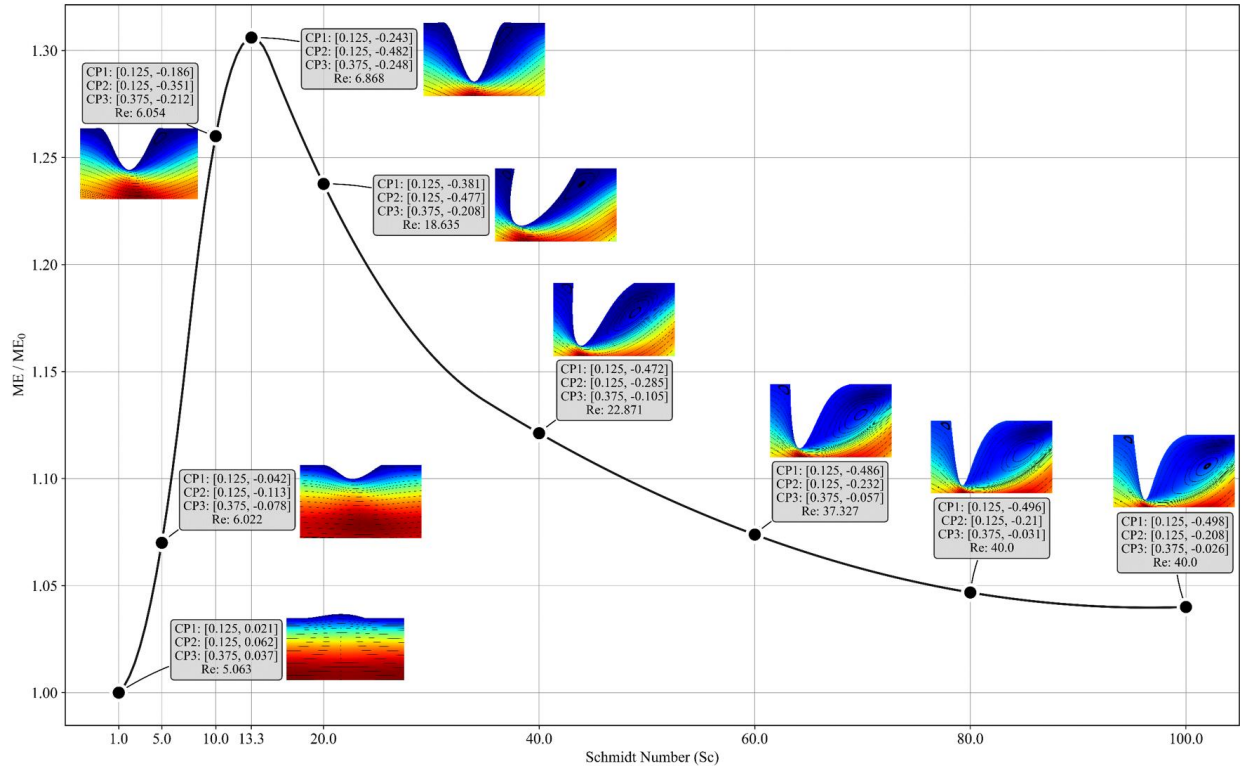

**Figure 9.** Optimal designs (with design parameters including the Reynolds number and spline control points) and their corresponding relative mixing efficiency for each Schmidt number.

For a more detailed analysis of the optimal configurations across different Schmidt numbers, Figure 10 illustrates the specific variations of the Reynolds number and spline parameters for these optimized cases. As evidenced by the figure, lower Reynolds numbers yield optimal solutions at low Schmidt numbers, whereas higher Reynolds numbers are optimal in the high Schmidt number regime. Furthermore, the trends in the spline control points reveal a distinct pattern: the value of CP1 generally decreases, while CP2 and CP3 initially decrease and subsequently increase. This specific progression of the spline parameters facilitates the formation of a symmetric, internal baffle up to a Schmidt number of 13.3, transitioning to an asymmetric baffle-shaped geometry beyond this point. The chart is constructed such that each vertical line represents the set of optimal design parameters for a specific Schmidt number. For further details and animated visualizations (GIF format) of the spline parameter evolution, please refer to the GitHub repository.

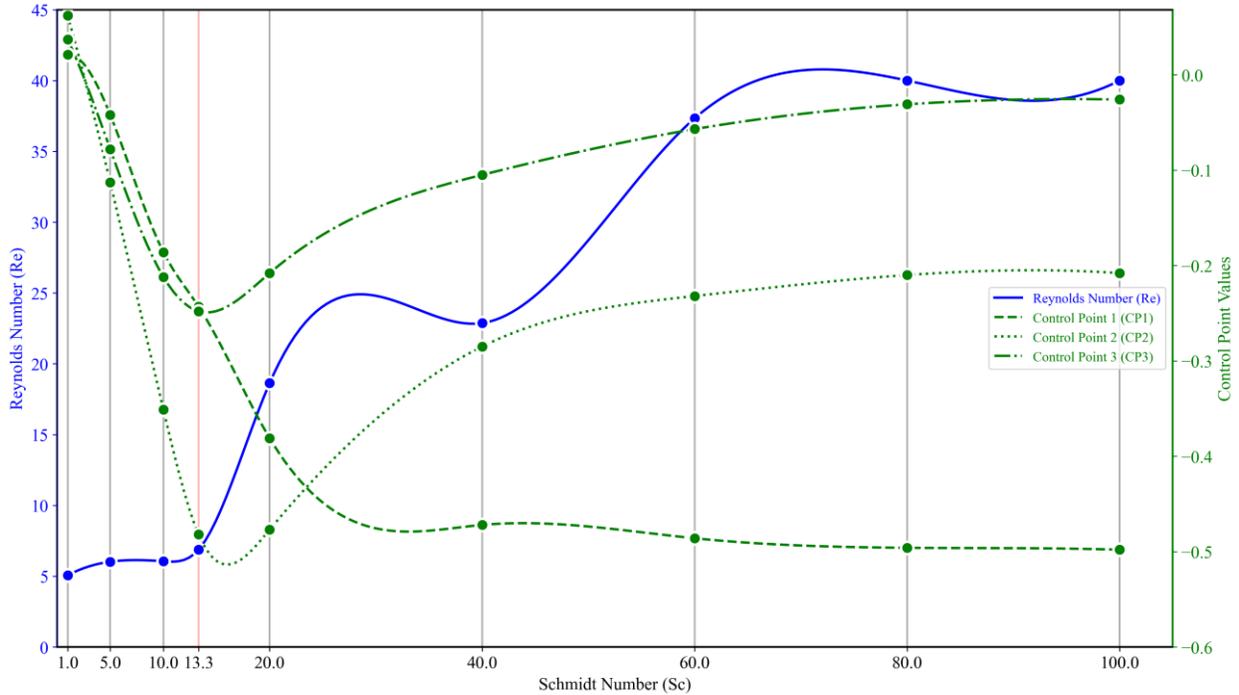

**Figure 10.** Parameter study of the optimal design configurations achieving maximum mixing efficiency across the Schmidt number range. (Note: Each vertical line in the plot represents the set of optimal design parameters for a specific Schmidt number).

Finally, to confirm the solutions obtained from the proposed optimization method, a parametric study was also conducted using a Genetic Algorithm (GA), which yielded its own set of optimal solutions. In the GA, an optimal solution must be found for each individual Schmidt number after the parametric analysis is complete [50]. Consequently, as shown in Figure 11, the computational cost of this process increases linearly with the number of Schmidt numbers considered. In contrast, the DRL-based agent employed in this study possesses learning capabilities. After an initial, more computationally intensive training phase, the trained agent can provide near-instantaneous optimal solutions for any number of input samples (i.e., Schmidt numbers). As the graph clearly indicates, the proposed DRL method becomes computationally more efficient than the GA when the number of samples exceeds 1,296. Moreover, unlike the Genetic Algorithm, which relies purely on stochastic search and requires repeated evaluations, the reinforcement learning (RL)-based framework exploits gradient-based updates and learns a generalized mapping between flow parameters and optimal designs. This enables the agent to generalize its learned policy to unseen Schmidt numbers without retraining, resulting in a significant reduction in computational cost for parametric studies. The proposed method also produces smoother and more physically consistent design transitions across parameter variations, which is crucial for manufacturable geometries and stable flow performance. Overall, the RL-based optimization demonstrates superior scalability, sample efficiency, and adaptability compared to population-based algorithms such as GA [51-52]. For detailed implementation of the GA and its hybrid integration with the proposed DRL framework, please refer to the GitHub repository.

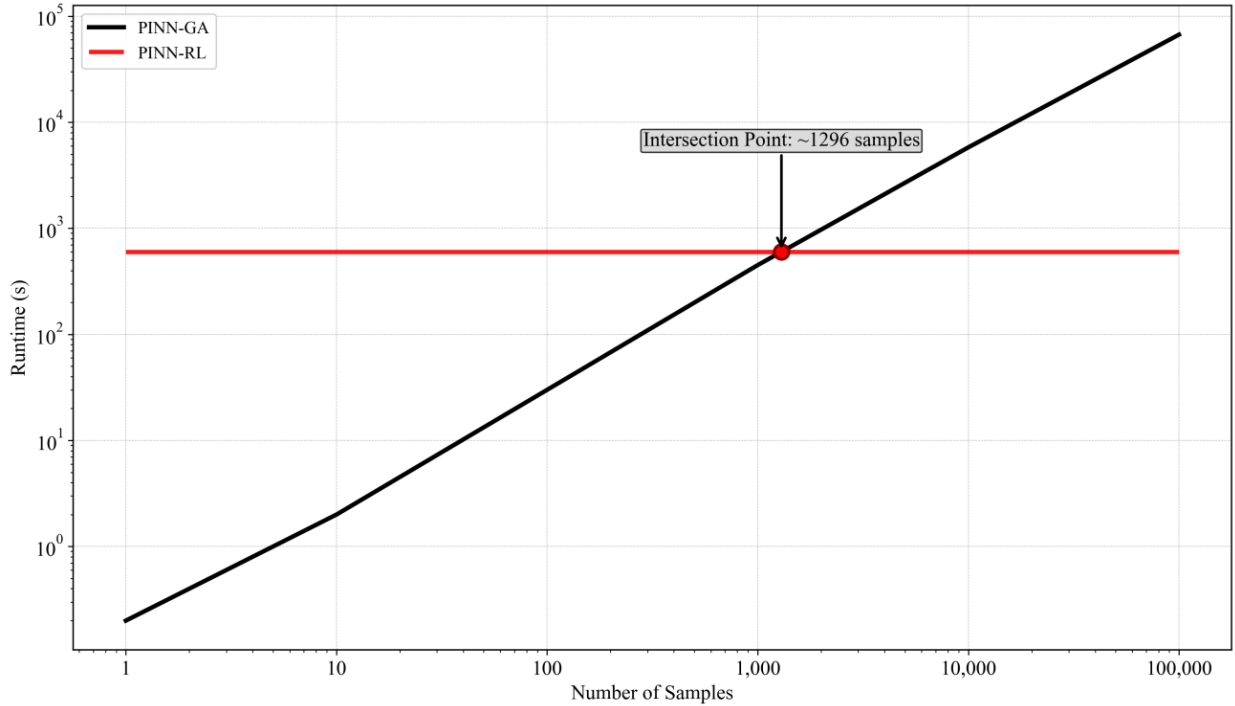

**Figure 11.** Computational time versus sample count for the PINN-RL method compared to the Genetic Algorithm (GA).

**Conclusion**

This study successfully established a novel Scientific Machine Learning (Sci-ML) framework for the generalized, efficient optimization of multi-parameter systems, demonstrated through the design of a passive micromixer. By integrating a parametrically trained Physics-Informed Neural Network (PINN) as a rapid simulation environment with a Generalized Deep Reinforcement Learning (GDRL) agent, we have addressed critical limitations of traditional optimization approaches. The primary achievement of this work is the development of a model that, after a single training phase, provides instantaneous optimal solutions across a wide range of physical conditions, eliminating the need for case-specific retraining or computationally expensive simulations.

The optimization process yielded physically insightful results, revealing a clear correlation between the optimal design and the dominant mixing mechanism. For Schmidt numbers up to 13.3, where molecular diffusion prevails, the optimal configuration consistently featured a symmetric, internal baffle. Beyond this transitional point, as convective mass transfer became dominant, the optimal geometry shifted towards an asymmetric internal baffle. This shift enhanced mixing through intensified vortices; however, the associated increase in pressure loss ultimately led to a net reduction in the relative mixing efficiency at higher Schmidt numbers, highlighting the critical trade-off between mixing enhancement and hydraulic cost. The framework consistently generated designs that outperformed the baseline case across the entire spectrum, with a peak performance gain of approximately 32% observed at a Schmidt number of 13.3.

The advantages of the proposed methodology were further underscored through a comparative analysis with a Genetic Algorithm (GA). The results demonstrated that while the initial training of the DRL agent is computationally intensive, its ability to provide near-instantaneous solutions for any number of subsequent queries makes it vastly more efficient for extensive parametric studies. The crossover point, beyond which our method outperforms the GA in computational speed, was found to be a sample size of 1,296, proving its superior scalability. In summary, this research presents a robust, efficient, and generalizable optimization paradigm that bridges scientific computing and machine learning, with significant potential for application to a broad class of complex, multi-parameter design problems in fluid dynamics and beyond. The proposed optimization framework demonstrates significant potential for generalization beyond the current application. It can be adapted to a wide range of other scientific and engineering domains, such as heat transfer, turbulence modeling, and any field requiring efficient exploration of high-dimensional parameter spaces.

Future research directions are multi-faceted. Firstly, the current methodology conditions the optimization on a single key parameter, the Schmidt number. A compelling extension would be to increase the dimensionality of the state space by incorporating multiple governing parameters simultaneously, thereby enabling the solution of more complex, multi-faceted design problems. Secondly, while the parametric PINN served as an efficient and high-fidelity environment in this study, the underlying reinforcement learning architecture is inherently compatible with real-world experimental setups. Replacing the PINN environment with real-time laboratory systems would represent a critical step towards bridging the gap between simulation and practical implementation, facilitating direct, real-time optimization of physical systems.

## CRediT authorship contribution statement

**Meraj Hassanzadeh:** Conceptualization, Methodology, Validation, Visualization, Software, Investigation, Writing the original draft.

**Ehsan Ghaderi:** Conceptualization, Methodology, Validation, Visualization, Software, Investigation, Writing the original draft.

**Mohamad Ali Bijarchi:** Review and Editing, Supervision.

**Siamak Kazemzadeh Hannani:** Review and Editing, Supervision.

## Acknowledgment

The authors gratefully acknowledge the assistance of AI-based language tools in improving the clarity and fluency of the English writing. All ideas, arguments, and scientific content are solely the responsibility of the authors.

## Statements and Declarations

No funding was received to assist with the preparation of this manuscript.

# Supplementary Section 1: Baffle Geometry Generation and Boundary Condition Implementation

## S1.1. Mathematical Formulation of Baffle Geometry

### S1.1.1 Control Points Definition

The baffle contour is defined by five control points in the Cartesian plane:

$P0 = (0.0, 0.0), P1 = (0.125, CP_1), P2 = (0.25, CP_2), P3 = (0.375, CP_3), P4 = (0.5, 0.0)$

In this formulation, the y-coordinates of the control points ($CP_n$) serve as the sole parametric variables. A diverse set of spline shapes can be generated by systematically perturbing these y-values, while the x-coordinates remain fixed to ensure uniform parameterization. A graphical example of this parameterization is provided in Figure S.1.

### S1.1.2 Cubic Spline Interpolation

The baffle geometry is constructed as a cubic spline passing through all control points. For each interval $[x_i, x_i + 1]$ between consecutive control points, the curve is described by a cubic polynomial:

$$S_i(x) = a_i + b_i(x - xi) + c_i(x - xi)^2 + d_i(x - xi)^3$$

where $i = 0,1,2,3$ denotes the segment index.

### S1.1.3 Spline Coefficient Determination

The coefficients $a_i, b_i, c_i, d_i$ are determined by enforcing the following conditions:

1. **Interpolation conditions**:

$S_i(x_i) = y_i \text{ and } S_i(x_i + 1) = y_i + 1$

2. **First derivative continuity** (C1):

$S_{i'}(x_i + 1) = S_{i'+1}(x_i + 1)$

3. **Second derivative continuity** (C2C2):

$S_{i''}(x_i + 1) = S_{i''+1}(x_i + 1)$

4. **Natural boundary conditions**:

$S_{0''}(x_0) = 0 \text{ and } S_{3''}(x_4) = 0$

This yields a tridiagonal system of equations that uniquely determines the spline coefficients.

## S1.2. Normal Vector Computation

### S1.2.1 Tangent Vector Derivation

The first derivative of the spline function provides the slope of the curve:

$$S_i'(x) = b_i + 2c_i(x - x_i) + 3d_i(x - x_i)^2$$

The tangent vector $T(x)$ at any point along the curve is:

$$T(x) = \begin{bmatrix} 1 \\ S'(x) \end{bmatrix}$$

### S1.2.2 Unit Tangent Vector

The unit tangent vector is obtained by normalization:

$$\hat{T}(x) = \frac{T(x)}{\| T(x) \|} = \frac{1}{1 + [S'(x)]^2} \begin{bmatrix} 1 \\ S'(x) \end{bmatrix}$$

### S1.2.3 Unit Normal Vector

The unit normal vector $\hat{n}(x)$ is obtained by rotating the unit tangent vector counterclockwise by 90°:

$$\hat{n}(x) = \frac{1}{1 + [S'(x)]^2} \begin{bmatrix} -S'(x) \\ 1 \end{bmatrix}$$

This normal vector points to the left of the direction of increasing $x$ along the curve.

## S1.3. Boundary Condition Implementation

### S1.3.1 No-Flux Boundary Condition

The mass conservation boundary condition at the baffle surface requires that the normal component of the mass flux vanishes:

$$J \cdot \hat{n} = 0$$

where J is the mass flux vector and $\hat{n}$ is the unit normal vector derived in Section S2.3. This concept is visualized in Figure S.1.

### S1.3.2 Physical Interpretation

This condition ensures that no mass crosses the baffle surface, making it an impermeable boundary. Mathematically, this constrains the velocity field such that its component normal to the baffle surface is zero at the boundary.

### S1.3.3 Discrete Implementation

For numerical implementation, the boundary condition is enforced at discrete sample points $x_k$ along the channel walls:

$$J(x_k, S(x_k)) \cdot \hat{n}(x_k) = 0 \quad for \quad k = 1, 2, \ldots, N$$

where the sample points are uniformly distributed along the x-domain of the baffle. This mathematical formulation guarantees consistent implementation of the no-flux boundary condition in computational simulations of mass transport around the baffle structures.

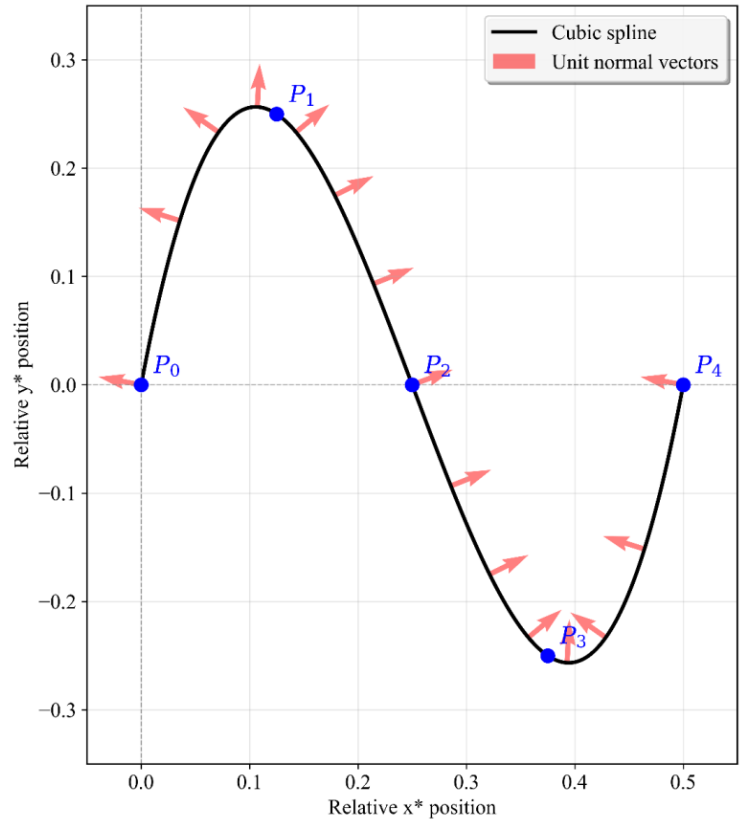

**Figure S1.** Baffle geometry generation via cubic spline interpolation. Control points (blue) define spline contours with analytically computed normal vectors (red) for boundary condition implementation. Lower panels illustrate geometric variations achieved through parametric y-coordinate modifications.

## Supplementary Section 2: PPO-PINN Algorithm Overview

The proposed algorithm combines Proximal Policy Optimization (PPO) with Physics-Informed Neural Networks (PINNs) to optimize mixing performance in microfluidic devices. The algorithm learns optimal control parameters that maximize mixing efficiency while minimizing pressure costs.

### S2.1. Mathematical Formulation

#### S2.1.1 State and Action Spaces

**State Space:**

- Schmidt number: $Sc \in [1, 100]$
- State vector: $s = [Sc]$

**Action Space:**

- Control parameters: $a = [cp_1, cp_2, cp_3, Re]$
- Physical constraints:
  - $cp^1, cp^2, cp^3 \in [-0.5, 0.5]$
  - $Re \in [5, 40]$

#### S2.1.2 Policy Representation

The actor network implements a Gaussian policy:

- Mean: $\mu_{\theta(s)} = \mu_{ActorNetwork}(s)$
- Standard deviation: $\sigma_{\theta(s)} = \exp(log\sigma_{ActorNetwork}(s))$
- Policy distribution: $\pi_{\theta(a|s)} = \mathcal{N}(a|\mu_{\theta(s)}, \sigma_{\theta(s)})$

### S2.2. PINN-Based Reward Function

#### S2.2.1 Field Evaluation

For each state-action pair $(s, a)$:

**Inlet Region Evaluation:**

- Input tensor: $X_{inlet} = [x, y, cp^1, cp^2, cp^3, Re, Sc]$
- Pressure field: $p = PINN_{pressure}(X_{inlet})$

**Outlet Region Evaluation:**

- Input tensor: $X_{outlet} = [x, y, cp_1, cp_2, cp_3, Re, Sc]$
- Concentration field: $c = PINN_{concentration}(X_{outlet})$

### S2.2.2 Performance Metrics

**Mixing Index:**

$$MI = 1 - \sqrt{\left[\left(\frac{1}{N}\right) \times \sum_{i=1}^{N} \left(\frac{c_i - 0.5}{0.5}\right)^2\right]}$$

**Pressure Cost:**

$$C_p = \left(\frac{1}{N}\right) \times \sum_{i=1}^{N} p_i$$

**Reward Function:**

$$r = \frac{\frac{MI}{MI_0}}{\frac{\sqrt[3]{C_p}}{\sqrt[3]{C_{p0}}}}$$

## S2.3. PPO Optimization

### S2.3.1 Advantage Estimation

**Value Function:**
$$V_{\varphi(s)} = \varphi_{CriticNetwork}(s)$$

**Advantage:**
$$A(s, a) = r(s, a) - V_{\varphi(s)}$$

**Normalized Advantage:**
$$\hat{A}(s, a) = \frac{(A(s, a) - \mu_A)}{(\sigma_A + \varepsilon)}$$

### S2.3.2 Policy Optimization Objective

**Probability Ratio:**
$$r_{t(\theta)} = \frac{\pi_{\theta(a_t|s_t)}}{\pi_{\theta_{old}(a_t|s_t)}}$$

**Clipped Objective:**
$$L^{CLIP(\theta)} = \mathbb{E}\left[\min\left(r_{t(\theta)\hat{A}_t}, clip(r_{t(\theta)}, 1-\varepsilon, 1+\varepsilon)\hat{A}_t\right)\right]$$

**Value Function Loss:**
$$L^{VF(\varphi)} = \mathbb{E}\left[\left(V_{\varphi(s_t)} - r_t\right)^2\right]$$

**Entropy Bonus:**
$$H(\pi_\theta) = \mathbb{E}\left[-\log \pi_{\theta(a|s)}\right]$$

**Total Loss:**
$$L^{TOTAL} = L^{CLIP(\theta)} - c^1 L^{VF(\varphi)} + c^2 H(\pi_\theta)$$

$$\text{where } c^1 = 1.0, c^2 = 0.01$$

## S2.4. Training Procedure

**Batch Processing**

For each episode:

1. Sample N Schmidt numbers: $Sc_i \sim \mathcal{U}(1, 100)$
2. Form state batch: $S = [Sc^1, Sc^2, \ldots, Sc_N]$
3. Generate actions: $A \sim \pi_{\theta(A|S)}$
4. Scale actions to physical bounds
5. Compute rewards: $R = [r(s^1, a^1), \ldots, r(s_N, a_N)]$

**Policy Update**

For K epochs:

1. Compute new policy $\pi_{\theta_{new}}$
2. Calculate probability ratios $r_{t(\theta)}$
3. Compute advantages $\hat{A}_t$
4. Update parameters:
$$\theta \leftarrow \theta - \alpha_\theta \nabla_\theta L^{TOTAL}$$
$$\varphi \leftarrow \varphi - \alpha_\varphi \nabla_\varphi L^{TOTAL}$$

**Hyperparameters**

- Discount factor: $\gamma = 0.99$
- Clipping parameter: $\varepsilon = 0.2$
- Optimization epochs: $K = 10$

- Learning rates: $\alpha_\theta = 3 \times 10^{-4}, \alpha_\varphi = 1 \times 10^{-3}$
- Batch size: $N = 64$
- Total episodes: $E = 100$

This formulation enables efficient optimization of mixing parameters across varying Schmidt numbers while respecting physical constraints and maximizing the mixing efficiency-to-pressure cost ratio.

To ensure reproducibility and facilitate further research, the full implementation of the PPO algorithm and the PINN environment has been made publicly available on [GitHub](#).